\documentclass[sn-mathphys-num]{sn-jnl}%

\usepackage{graphicx}%
\usepackage{multirow}%
\usepackage{amsmath,amssymb,amsfonts}%
\usepackage{amsthm}%
\usepackage{mathrsfs}%
\usepackage[title]{appendix}%
\usepackage{xcolor}%
\usepackage{textcomp}%
\usepackage{manyfoot}%
\usepackage{tabularx} %
\usepackage{manyfoot}%
\usepackage{booktabs}%
\usepackage{caption}%
\usepackage{algorithm}%
\usepackage{algorithmic}
\usepackage{float}%
\usepackage{listings}%
\usepackage{bm}
\usepackage{subfigure}
\usepackage{epigraph}

\usepackage[capitalise, noabbrev]{cleveref}
\definecolor{mydarkblue}{rgb}{0.0, 0.0, 0.5}

\newcommand{\method}{D-SINK}
\newcommand{\Method}{Dual-granularity Sinkhorn Distillation}

\theoremstyle{thmstyleone}%

\theoremstyle{thmstyletwo}%

\theoremstyle{thmstylethree}%

\begin{document}

\title{Dual-granularity Sinkhorn Distillation for Enhanced Learning from Long-tailed Noisy Data}

\author[1]{\fnm{Feng} \sur{Hong}}\email{feng.hong@sjtu.edu.cn}\equalcont{These authors contributed equally to this work.}

\author[1]{\fnm{Yu} \sur{Huang}}\email{hy010119@sjtu.edu.cn}\equalcont{These authors contributed equally to this work.}

\author[1]{\fnm{Zihua} \sur{Zhao}}\email{sjtuszzh@sjtu.edu.cn}

\author[1]{\fnm{Zhihan} \sur{Zhou}}\email{zhihanzhou@sjtu.edu.cn}

\author*[1]{\fnm{Jiangchao} \sur{Yao}}\email{Sunarker@sjtu.edu.cn}

\author[3]{\fnm{Dongsheng} \sur{Li}}\email{ dongsheng.li@microsoft.com}

\author[2]{\fnm{Ya} \sur{Zhang}}\email{ya\_zhang@sjtu.edu.cn}

\author*[2]{\fnm{Yanfeng} \sur{Wang}}\email{wangyanfeng622@sjtu.edu.cn}

\affil[1]{\orgdiv{Cooperative Medianet Innovation Center}, \orgname{Shanghai Jiao Tong University}, \orgaddress{ \city{Shanghai}, \postcode{200240},  \country{China}}}

\affil[2]{\orgdiv{School of Artificial Intelligence}, \orgname{Shanghai Jiao Tong University}, \orgaddress{ \city{Shanghai}, \postcode{200230},  \country{China}}}

\affil[3]{\orgname{Microsoft Research Asia}, \orgaddress{\city{Shanghai}, \postcode{200232}, \country{China}}}

\abstract{
Real-world datasets for deep learning frequently suffer from the co-occurring challenges of class imbalance and label noise, hindering model performance. 
While methods exist for each issue, effectively combining them is non-trivial, as distinguishing genuine tail samples from noisy data proves difficult, often leading to conflicting optimization strategies. 
This paper presents a novel perspective: instead of primarily developing new complex techniques from scratch, we explore synergistically leveraging well-established, individually `weak' auxiliary models—specialized for tackling either class imbalance or label noise but not both. 
This view is motivated by the insight that class imbalance (a distributional-level concern) and label noise (a sample-level concern) operate at different granularities, suggesting that robustness mechanisms for each can in principle offer complementary strengths without conflict.
We propose {\Method} ({\method}), a novel framework that enhances dual robustness by distilling and integrating complementary insights from such `weak', single-purpose auxiliary models. 
Specifically, {\method} uses an optimal transport-optimized surrogate label allocation to align the target model's sample-level predictions with a noise-robust auxiliary and its class distributions with an imbalance-robust one.
Extensive experiments on benchmark datasets demonstrate that {\method} significantly improves robustness and achieves strong empirical performance in learning from long-tailed noisy data.
}


\maketitle

\section{Introduction}\label{Sec: Intro}
\begin{quote}
\itshape 
``No one can whistle a symphony. It takes a whole orchestra to play it.''
\nopagebreak 
\raggedleft 
--- \textsc{H.E. Luccock} 
\end{quote}

Deep learning has witnessed exponential growth across diverse fields like computer vision~\citep{DBLP:conf/cvpr/HeZRS16,MoA-VR,VARFVV,DBLP:conf/iccv/CaronTMJMBJ21}, natural language processing~\citep{DBLP:conf/naacl/DevlinCLT19,DBLP:journals/corr/abs-1907-11692,zhou2025learning}, and automated healthcare~\citep{DBLP:journals/mia/LitjensKBSCGLGS17,DBLP:journals/tmi/DaiZHYZW24,DBLP:journals/mia/HolsteZWJLZYKMTJPRHVYKSKCLSSWP24}. This revolution is largely fueled by the availability of large-scale, high-quality datasets~\citep{DBLP:conf/cvpr/DengDSLL009}. However, acquiring such ideal training data in real-world applications is often a major challenge. Two primary issues consistently emerge: 1) Class imbalance, a consequence of the real world's inherently long-tailed distributions~\citep{DBLP:journals/pami/ZhangKHYF23,cao2025analytical}, where some classes are vastly overrepresented while others are scarce; and 2) Label noise, where samples suffer from incorrect annotations due to inherent ambiguity, annotator error, or flaws in automated labeling~\citep{DBLP:journals/tnn/SongKPSL23}. Prevalent large-scale datasets, such as WebVision~\citep{DBLP:journals/corr/abs-1708-02862}, clearly demonstrate this dual challenge, exhibiting both significant class imbalance and pervasive label noise.

For decades, researchers have tackled these issues, though largely in isolation. Long-tailed learning techniques—spanning class re-weighting~\citep{DBLP:conf/cvpr/CuiJLSB19,DBLP:conf/iclr/0004YL0T0W24,DBLP:conf/nips/Luo0Y00024},
re-sampling~\citep{DBLP:journals/jair/ChawlaBHK02,DBLP:conf/iclr/AhnKY23}, logit adjustment~\citep{DBLP:conf/iclr/MenonJRJVK21,DBLP:conf/iclr/0004Y00W23,DBLP:conf/nips/RenYSMZYL20}, and decoupled training~\citep{DBLP:conf/iclr/KangXRYGFK20,DBLP:conf/nips/0002Y0ZHW23}—aim to counteract biases from skewed class distributions. Concurrently, noisy label learning has matured with methods like the small loss trick~\citep{DBLP:conf/nips/HanYYNXHTS18,DBLP:conf/iclr/LiSH20}, label refinement~\citep{DBLP:conf/cvpr/YaoSZS00T21,DBLP:journals/corr/abs-2507-12998,DBLP:conf/cvpr/KarimRRMS22}, and regularization~\citep{DBLP:conf/icml/00030YYXTS20} to mitigate label corruption. However, when faced with real-world data where both challenges co-exist, simply ``bolting together'' these solutions proves ineffective~\citep{DBLP:journals/corr/abs-2108-11569}. This is largely because the presence of both extreme class imbalance and label noise makes it very hard to distinguish genuine tail samples from noisy instances. This confusion can create direct conflicts: methods designed to boost tail classes might unintentionally strengthen the impact of noisy labels within them, while techniques aimed at correcting noisy labels might wrongly remove valuable tail samples. Recognizing this gap, recent efforts~\citep{DBLP:journals/corr/abs-2108-11569,DBLP:conf/aaai/HuangBZBW22,DBLP:conf/iccv/LuZHCW23} have focused on developing specialized techniques from the ground up for long-tailed noisy label learning, even though the wide range of methods for tackling each of these data imperfections—class imbalance or label noise—when addressed in isolation is already considerably well-developed.

Our research approaches this problem from an entirely different, even counter-intuitive angle: \emph{``Can we achieve significant breakthroughs in learning from long-tailed noisy data by intelligently combining insights from `weak', single-purpose auxiliary models – designed only for long-tailed learning or only for noisy-label learning?''} These auxiliary models, while effective in their specific area, would typically be considered weak on their own for the combined, complex task. The motivation for this question lies in the distinct natures of these imperfections: class imbalance is a distributional concern, while label noise affects individual samples. 
We hypothesize that these distinct robustness mechanisms, despite the simplicity or `weakness' of the individual auxiliary models when facing the dual problem, can offer complementary strengths. If harmonized, they could lead to a powerful solution without inherent conflict. 

This paper provides a comprehensive empirical validation of this hypothesis, demonstrating through our proposed framework and extensive experiments that the answer to our guiding question is a clear `yes'. 
We introduce {\Method} ({\method}), a novel framework that achieves remarkable robustness to both class imbalance and label noise by skillfully distilling and integrating knowledge from multiple auxiliary models, each possessing only a single type of robustness. 
Crucially, {\method} does not require these auxiliary models to be strong or complex; instead, it leverages their focused, though `weak' in isolation, expertise.
Specifically, during the training of the target model, we introduce a surrogate label allocation to ensure alignment of the target model's outputs with those of the noise-robust auxiliary model at the sample level, while aligning distributions with the imbalance-robust auxiliary model. The surrogate label allocation can be efficiently solved through an efficient optimal-transport optimization process, allowing the target model to leverage auxiliary models effectively, thereby enhancing its robustness to both imbalance and noise.  We summarize the contributions as follows:

\begin{enumerate}
    \item \textbf{A New Perspective:} 
    We introduce and validate a novel perspective: that significant advancements in long-tailed noisy label learning can be achieved by synergistically combining insights from multiple `weak' auxiliary models, each specialized for only a single aspect (class imbalance or label noise).
    \item \textbf{A Novel Method:} We propose {\method}, a new framework that effectively distills and integrates complementary knowledge from multiple single-robustness auxiliary models into a target model. {\method} employs a surrogate label allocation design, optimized via an efficient optimal-transport process, to align the target model with noise-robust auxiliary models at the sample level and with imbalance-robust auxiliary models at the distribution level.
    \item \textbf{Strong Empirical Performance:} We conduct extensive experiments on benchmark datasets featuring diverse noise patterns, varying noise ratios and imbalance ratios. These evaluations show that {\method} yields substantial performance gains, highlighting the surprising effectiveness of its strategy: leveraging and harmonizing `weak', individually focused auxiliary models to navigate the complexities of the dual-challenge scenario.
\end{enumerate}

\section{Related Work}\label{Sec: Related Work}
\subsection{Noisy Label Learning}\label{SubSec: Related Work: noisy}

Noisy Label Learning (NLL) focuses on mitigating the prevalent label corruption issue encountered in modern web-scraped datasets. Prevailing approaches in NLL encompass various strategies, which can be broadly categorized into transition matrix estimation, robust regularization and sample selection. Transition matrix estimation, initially proposed by~\citep{DBLP:journals/jmlr/RooyenW17}, tackles this problem by estimating the probabilities of label flipping. Volume minimization network (VolMinNet)~\citep{DBLP:journals/pmlr/li2021provably} further advances this concept by stabilizing the estimation of the noise transition matrix.
Robust regularization techniques are employed to prevent models from overfitting to noisy labels. For instance, survey recipes for building
 reliable and robust deep networks (SURE)~\citep{DBLP:journals/corr/abs-2403-00543} combines entropy maximization with Sharpness-Aware Minimization~\citep{DBLP:conf/iclr/ForetKMN21} to improve generalization, while stochastic integrated gradient underweighted ascent (SIGUA)~\citep{DBLP:conf/icml/00030YYXTS20} applies regularization directly to the labels themselves. Sample selection methods have demonstrated significant effectiveness and become the mainstream in dealing with noisy labels.  Learning with noisy labels as semi-supervised learning (DivideMix)~\citep{DBLP:conf/iclr/LiSH20} uses a Gaussian Mixture Model to separate clean and noisy samples, subsequently employing semi-supervised training. Combating label noise through uniform selection and contrastive learning (UNICON)~\citep{DBLP:conf/cvpr/KarimRRMS22} integrates sample selection with contrastive learning to more effectively distinguish noisy samples, particularly under high noise ratios.

\subsection{Long-Tailed Learning}\label{SubSec: Related Work: long-tailed}

Long-Tailed Learning (LTL) addresses the challenges of imbalanced data distributions, where research predominantly falls into four categories including class re-weighting, data re-sampling, logit adjustment and decoupled training. Class re-weighting strategies aim to mitigate imbalance by adjusting loss contributions, exemplified by Class-Balanced Loss~\citep{DBLP:conf/cvpr/CuiJLSB19}, which introduces ``effective number'' to approximate exact per-class sample amount, and Influence-Balanced Loss~\citep{DBLP:conf/iccv/ParkLJ021}, which proposes a robust loss function to re-weight tail class samples. Data re-sampling strategies techniques directly alter the training data distribution, for instance, through over-sampling minority or under-sampling majority classes as demonstrated by learning with a Dirichlet mixture of the
 experts (DirMixE)~\citep{DBLP:journals/corr/abs-2405-07780} and curriculum of data augmentation (CUDA)~\citep{DBLP:conf/iclr/AhnKY23}. Another line of work, Logit adjustment~\citep{DBLP:conf/iclr/MenonJRJVK21}, seeks to solve class imbalance by modifying model output logits based on label frequencies. Adversarial robustness under long-tailed distribution (RoBal)~\citep{DBLP:conf/cvpr/Wu0H0L21} refine this via applying a post-processing adjustment to cosine classifiers. Decoupled training decouples the learning procedure into representation learning and classifier training. For instance,  $k$-positive contrastive learning (KCL)~\citep{DBLP:conf/iclr/KangLXYF21} develops a $k$-positive contrastive loss to learn a more class-balanced and class-discriminative feature space.  Distribution alignment (DisAlign)~\citep{DBLP:conf/cvpr/ZhangLY0S21} innovated the classifier training with adaptive logit adjustment.

\subsection{Long-Tailed Noisy Label Learning}\label{SubSec: Related Work: ImbNoise}

While extensive research has been conducted in either Long-Tailed Learning (LTL) or Noisy Label Learning (NLL) independently, methods designed for one scenario often prove ineffective when both challenges co-occur. This inefficacy stems from the inherent conflicts, that LTL may inadvertently amplify the negative effects of noisy labels when up-weighting the tailed classes. Conversely, NLL methods risk mis-classifying scarce tail-class samples as noisy data. Consequently, specialized Long-Tailed Noisy Label Learning (LT-NLL) approaches have emerged to jointly address these intertwined challenges. For instance, two-stage bi-dimensional
 sample selection (TABASCO)~\citep{DBLP:conf/iccv/LuZHCW23} focuses on label noise within intrinsically long-tailed data distributions. Heteroskedastic adaptive regularization (HAR)~\citep{DBLP:conf/iclr/CaoCLAGM21} and robust long-tailed learning under label noise (RoLT)~\citep{DBLP:journals/corr/abs-2108-11569} leverage adaptive regularization, prototypical error detection, and semi-supervised techniques to mitigate label noise under long-tailed distributions. Fairness regularizer (FR)~\citep{DBLP:journals/corr/abs-2303-12291} introduces a fairness regularizer aimed at improving performance on tail subpopulations. Meta-weight-net (MW-Net)~\citep{DBLP:conf/nips/ShuXY0ZXM19} further employs meta-learning to dynamically re-weight training samples for enhancing generalization. Representation calibration (RCAL)~\citep{DBLP:conf/iccv/ZhangZYY023} addresses learning with noisy labels on long-tailed data by leveraging contrastive learning to obtain robust representations, and calibrating their distributions to recover clean data characteristics for improved classifier training. However, these studies often propose methods designed from the ground up, rather than explicitly leveraging or adapting well-established NLL or LTL strategies. As a result, further improvements are necessary to advance in this field.

 \subsection{Sinkhorn Distillation}\label{SubSec: Related Work: Sinkhorn Distillation}
 Optimal Transport (OT) provides a principled framework for comparing probability distributions by finding the minimal cost to transform one into another~\citep{monge1781mémoire,santambrogio2015optimal}. While traditionally used in mathematics and graphics, its application has recently expanded into deep learning, particularly for tasks requiring robust distribution alignment. The development of the Sinkhorn-Knopp algorithm~\citep{DBLP:conf/nips/Cuturi13} has been pivotal, offering an efficient, differentiable, and scalable approximation for solving entropically regularized OT problems. This has enabled the emergence of ``Sinkhorn Distillation'', a paradigm that leverages OT for knowledge transfer. A key advantage of OT-based distillation over conventional methods, which typically rely on Kullback-Leibler (KL) divergence, is its ability to incorporate the metric structure of the label space. Entropy-based losses treat all misclassifications equally, whereas OT can account for the fact that some errors are ``less wrong'' than others. This principle has been effectively applied in various contexts, for instance, the KNOT (knowledge distillation using optimal transport) framework~\citep{DBLP:conf/coling/BhardwajVP22} leverages OT to distill semantic knowledge from multiple teacher models in a federated learning setting by preserving the metric structure of the label space. The influence of Sinkhorn-based methods also extends to self-supervised learning, where approaches like SwAV (swapping assignments between
 multiple views of the same image)~\citep{DBLP:conf/nips/CaronMMGBJ20} use the Sinkhorn algorithm to perform large-scale online clustering, enabling the learning of powerful invariant features.

\section{Method}\label{Sec: Method}

\subsection{Problem Formulation}\label{subsec: Problem Formulation}
Let $\mathcal{X}$ be the input space and $\mathcal{Y} = \{1,2,\ldots,C\}$ be the label space, where $C$ is the class number. A long-tailed noisy-label training set can be denoted as $\mathcal{D} = \{\bm{x}_i,\hat{y}_i,y_i\}_{i=1}^N \in (\mathcal{X},\mathcal{Y},\mathcal{Y})^N$, where $\hat{y}_i\in \mathcal{Y}$ is the observed label of the sample $\mathbf{x}_i$, and its ground truth $y_i\in \mathcal{Y}$ is invisible. (Long-tailed) The sample number $N_c$ of each class $c \in \mathcal{Y}$ in the descending order exhibits a long-tailed distribution. The imbalance ratio is defined as $\mathrm{IR} = \frac{\max_{c\in\mathcal{Y}}N_c}{\min_{c\in\mathcal{Y}}N_c}$. We consider a dataset highly imbalanced when this ratio is significantly greater than one (\emph{i.e.}, $\mathrm{IR}\gg 1$). (Noisy-labeled) Certain samples within the dataset contain incorrect labels. The noise ratio is defined as the ratio of mislabeled data in the training set $\mathrm{NR} = \frac{1}{N}\sum_{i=1}^{N}\mathbf{1}(\hat{y}_i\neq y_i)$, where $\mathbf{1}(\cdot)$ denotes the indicator function, with value $1$ when $\cdot$ is true and $0$ otherwise. For evaluation, a class-balanced test set $\mathcal{D}_{\text{test}}$ with clean labels is used. The goal of learning from long-tailed noisy data is to learn a model $f: \mathcal{X} \rightarrow p(\mathcal{Y})$ on $\mathcal{D}$ that minimizes the error rate on $\mathcal{D}_{\text{test}}$.

\subsection{{\Method}}

To address the dual challenges of class imbalance and label noise defined in Section 3.1, we start from a core insight: these two issues operate at different granularities. Class imbalance is a dataset-level, distributional property, whereas label noise is an instance-level corruption. 
This distinction enlightens us to consider whether we can explore a divide-and-conquer mechanism to synergistically combine the strengths of specialized methods for two areas.
Intuitively, let $f_L$  and  $f_N$ be auxiliary models, trained on the dataset  $\mathcal{D}$  using a long-tailed learning algorithm  $A_L$  and a noisy-label learning algorithm  $A_N$, respectively. We seek to leverage them to enhance the target model $f$ for learning from long-tailed noisy data, ensuring  $f$  achieves robust performance against both class imbalance and label noise.

A natural approach is to employ knowledge distillation from  $f_L$  and $f_N$ to guide the training of  $f$, formulating it as a multi-teacher distillation problem. Existing methods often use a weighted aggregation strategy for multiple teachers~\citep{DBLP:journals/ijon/LiuZW20,DBLP:conf/iclr/TanRHQZL19,DBLP:conf/wacv/PhamHD23}. For instance, with equal weighting in the loss function, the distillation loss can be defined as:

\begin{equation}
    \mathcal{L}_{\text{Dist}} = \sum_{i=1}^N \left[D_{\mathrm{KL}}(f_L(\bm{x}_i)\|f(\bm{x}_i)) + D_{\mathrm{KL}}(f_N(\bm{x}_i)\|f(\bm{x}_i))\right]
    \label{eq: naive distillation}
\end{equation}

However, while $f_L$ and $f_N$ individually exhibit robustness against long-tailed distribution and label noise, respectively, they inherently conflict with each other, as highlighted in many studies~\citep{DBLP:journals/corr/abs-2303-12291,DBLP:conf/iclr/CaoCLAGM21,DBLP:conf/eccv/YiT0LZ22,DBLP:conf/iccv/LuZHCW23}. For instance, for a difficult sample with a large loss, $f_L$  tends to treat it as a tail class sample, increasing the learning focus on that sample. In contrast, $f_N$ is likely to treat it as a noisy sample and may reduce or abandon learning from that sample. The direct weighted multi-teacher distillation approach, as shown in \cref{eq: naive distillation}, cannot resolve this inherent conflict and may lead to suboptimal results (as shown in \cref{tab:ablation study}).

\noindent\textbf{Dual-granularity Distillation.} 
The key insight behind {\method} is to leverage these auxiliary models at different granularities, reflecting their distinct robustness mechanisms: class imbalance is a dataset-level distributional property, while label noise pertains to individual samples. We propose that $f_L$ should guide the target model $f$ at the distributional level, while $f_N$ should provide guidance at the sample level.

To achieve this, we introduce a set of geometric proxy labels  $Q = [\textbf{q}_1, \textbf{q}_2, \ldots, \textbf{q}_N] \in \mathbb{R}^{C \times N}$. These proxy labels $\textbf{q}_i \in \Delta^C$ (where $\Delta^C$ is the $C$-dimensional probability simplex) for each sample $\bm{x}_i$ act as dynamic, refined targets for training $f$. They are determined by simultaneously satisfying two conditions:
\begin{enumerate}
\item At the sample level, $\textbf{q}_i$ should be close to the prediction of the noise-robust model $f_N(\bm{x}_i)$.
\item At the distribution level, the aggregated proxy labels $\sum_{i=1}^N \textbf{q}_i$ should match the overall class distribution predicted by the imbalance-robust model $\sum_{i}^N f_L(\bm{x}_i)$.
\end{enumerate}
By learning from these geometric proxy labels $Q$, the target model $f$ can simultaneously achieve strong robustness to both noisy labels and long-tailed distributions. Thus, the proposed distillation loss function can be defined as:
\begin{equation}
\begin{aligned}\label{Eq: loss}
    \mathcal{L}_{\mathrm{\method}} = &\frac{1}{N}\sum_{i=1}^N \left[\underbrace{D_{\mathrm{KL}}(\textbf{q}_i\|f_N(\bm{x}_i))}_{\text{\textcircled{1} Sample-level alignment of $Q$ and $f_N$}} + \underbrace{D_{\mathrm{KL}}(\textbf{q}_i\|f(\bm{x}_i))}_{\text{\textcircled{3} Learning $f$ from $Q$}}\right],\\ 
    &\text{ s.t., } \underbrace{Q \cdot \mathbf{1}_N = \sum_{i}^N f_L(\bm{x}_i), Q^\top \cdot \mathbf{1}_C = \mathbf{1}_N}_{\text{\textcircled{2} Distribution alignment of $Q$ and $f_L$}}. 
\end{aligned}
\end{equation}

Here, $\mathbf{1}_N$ and $\mathbf{1}_C$ denote column vectors of ones with dimensions $N$ and $C$, respectively. Each term in the loss function and constraint plays a distinct role in guiding the learning of the target model $f$:

\begin{itemize}
    \item \textcircled{1} \textbf{Sample-level alignment}: This term enforces that the geometric proxy labels $\textbf{q}_i$ approximate the predictions of the noise-robust model $f_N$ on a per-sample basis. It encourages $Q$ to capture the local semantics of each input $\bm{x}_i$ in a way that is resilient to label noise.
    
    \item \textcircled{2} \textbf{Distribution-level alignment}: This constraint regularizes the global behavior of $Q$ by aligning its aggregated predictions with those of the long-tailed-aware model $f_L$. Specifically, the first part $Q \cdot \mathbf{1}_N = \sum_{i=1}^N f_L(\bm{x}_i)$ ensures that the class-wise distribution of $Q$ matches that of $f_L$, while the second part $Q^\top \cdot \mathbf{1}_C = \mathbf{1}_N$ guarantees that each column of $Q$ forms a valid probability distribution.
    
    \item \textcircled{3} \textbf{Knowledge distillation to $f$}: This term encourages the target model $f$ to learn from the structured supervision signal $Q$. By minimizing the KL divergence between $\textbf{q}_i$ and $f(\bm{x}_i)$, the model $f$ internalizes the robustness and distribution-awareness encoded in $Q$, enabling better generalization.
\end{itemize}

Together, these components enable the target model $f$ to achieve robustness against both label noise and class imbalance, with $Q$ serving as a principled and trainable intermediate supervision.

\noindent\textbf{Overall Objective.} We leverage $\mathcal{L}_{\mathrm{\method}}$ from \cref{Eq: loss} to incorporate well-established methods from long-tailed learning and noisy-label learning, enhancing the robustness of the target model $f$ in both aspects. The overall objective of our method is defined as follows:
\begin{equation}
    \min_{Q,f} \mathcal{L}_{\mathrm{Overall}} = \mathcal{L}_{\mathrm{Base}} + \alpha \mathcal{L}_{\mathrm{\method}}, \label{Eq: objective}
\end{equation}
where $\mathcal{L}_{\mathrm{Base}}$  represents the basic loss function used to learn classification from the training set $\mathcal{D}$, and $\alpha$ is a hyperparameter that controls the weight of  $\mathcal{L}_{\mathrm{\method}}$. Here we can use the most commonly adopted cross-entropy loss as $\mathcal{L}_{\mathrm{Base}} = \frac{1}{N}\sum_{i=1}^{N} -\hat{y}_i\cdot\log f(\bm{x}_i)$. 
Alternatively, we can choose variants adapted for imbalanced noisy datasets, \emph{e.g.,} \citet{DBLP:journals/corr/abs-2108-11569, DBLP:conf/iccv/LuZHCW23}. 

\noindent\textbf{Bi-level Optimization.} 
The objective in \cref{Eq: objective} involves the simultaneous optimization of both  $f$ and $Q$, along with the constraints imposed on $Q$. This complexity makes it challenging to directly apply traditional gradient descent methods like SGD for optimization. Thus we optimize \cref{Eq: objective} in a bi-level style. 
\begin{itemize}
\item \textbf{Inner Loop (Optimizing $f$):} With $Q$ fixed, we update the parameters of the target model $f$ by minimizing $\min_f \mathcal{L}_{\mathrm{Base}} + \alpha \mathcal{L}_{\mathrm{\method}}$ using gradient descent (e.g., SGD and Adam).
\item \textbf{Outer Loop (Optimizing $Q$):} With $f$ fixed, we optimize for $Q$ by minimizing $\min_Q \mathcal{L}_{\mathrm{\method}}$ subject to its constraints. This subproblem can be elegantly reformulated as an optimal transport (OT) problem~\citep{monge1781mémoire,santambrogio2015optimal}. This formulation is particularly apt as OT naturally handles matching distributions under cost constraints, aligning with our goal of finding proxy labels $Q$ that mediate between $f_N$'s predictions (influencing the cost) and $f_L$'s marginal distribution (as a constraint).
\end{itemize}
Specifically, for the \textbf{Outer Loop}, we can reformulate the optimization of $Q$ in the outer loop as an optimal transport problem~\citep{monge1781mémoire,santambrogio2015optimal}, as follows:
\begin{equation}
\begin{aligned}\label{Eq: OT}
    \min_Q\mathcal{L}_{\mathrm{\method}} = &\min_Q\sum_{i=1}^N \textbf{q}_i\cdot(2\log \textbf{q}_i-\log f_N(\bm{x}_i) - \log f(\bm{x}_i))\\
    =&\min_Q \langle Q,P\rangle + 2 \sum_{i=1}^N \textbf{q}_i\cdot\log \textbf{q}_i,\\
    \text{ s.t., } &Q \cdot \mathbf{1}_N = \sum_{i=1}^N f_L(\bm{x}_i), Q^\top \cdot \mathbf{1}_C = \mathbf{1}_N,
\end{aligned}
\end{equation}
where $P = [p_1,\ldots,p_N]$, $p_i = -\log f_N(\bm{x}_i) - \log f(\bm{x}_i)$, and  $\langle \cdot, \cdot \rangle$  represents the matrix inner product, Here, $\log$ operates element-wise on the vector. \cref{Eq: OT} can be interpreted as an entropy-regularized optimal transport problem~\citep{DBLP:conf/nips/Cuturi13}, where $Q$ is the transport matrix, $P$ is the cost matrix with the entropic regularization term $2 \sum_{i=1}^N \textbf{q}_i \cdot\log \textbf{q}_i$, where ``2'' comes from the two KL divergence terms in \cref{Eq: loss}, each contributing one entropy component. 

To solve \cref{Eq: OT}, we employ the well-known Sinkhorn-Knopp algorithm~\citep{DBLP:conf/nips/Cuturi13} for efficient optimization. Formally, we define a matrix $M = e^{-\frac{P}{2}} \in \mathbb{R}^{C \times N}$, where the exponential is applied element-wise to the matrix, and obtain the solution of $Q$ as:
\begin{equation}\label{eq: sinkhorn}
\begin{aligned}
    Q &= N\cdot \mathrm{diag}(u) M \mathrm{diag}(v),\\
    \text{with iteratively updated \ \ } &u \leftarrow \frac{\sum_{i=1}^N f_L(\bm{x}_i)}{N}\ ./ \ (M v), v \leftarrow \frac{\mathbf{1}_N}{N}\ ./ \ (M^T u),
\end{aligned}
\end{equation}
where $u \in \mathbb{R}^C$ and $v \in\mathbb{R}^N$ are scaling coefficients
vectors. The operator $./$ denotes element-wise division, and $\mathrm{diag}(\cdot)$ represents the construction of a diagonal matrix from a vector. 
Derivation of the Sinkhorn algorithm can be referred to \cref{secA1}.

\subsection{Algorithmic Procedure and Complexity}
\begin{table}[!t] 
\centering
\caption{Training Time Comparison (seconds/batch): D-SINK vs. Standard Training, and computational overhead of {\method} across benchmark datasets. Overhead represents the difference.}
\label{tab:training_time_overhead} %
\begin{tabular}{lcccc}
\toprule
Method     & CIFAR-10 & CIFAR-100 & Red Mini-ImageNet & Clothing1M \\
\midrule
Standard Training & 1.271    & 1.274     & 3.888            & 3.951      \\
{\method}            & 1.426    & 1.418     & 4.317            & 4.263      \\
Overhead          & 0.155    & 0.144     & 0.429            & 0.312      \\
\bottomrule
\end{tabular}
\end{table}
We present the complete procedure of our method {\method} in \cref{algorithm:method}. Note that, since modern machine learning models are trained using the mini-batch approach, the optimization of proxy labels $Q$ and the computation of predictive distribution statistics are also performed in a mini-batch manner. Note that {\method} only use $f$ and do not need auxiliary models during testing or inference.

\noindent\textbf{Complexity Analysis.} Here, we analyze the additional complexity introduced by {\method} during training. Considering the time complexity, let the mini-batch size be $N_B$ and the number of model parameters be $P_\text{param}$. The time complexity of standard training is  $\mathcal{O}(N_BP_\text{param})$. {\method} introduces additional overhead primarily in the loop from lines 8 to 11 of \cref{algorithm:method}, with a time complexity of  $\mathcal{O}(N_BTC)$, where $T$ represents the number of iterations and $C$ denotes the number of classes. In implementation, we set $T = 50$, as it is sufficient for stable convergence of the iterations. Considering that modern machine learning models typically have parameter counts on the order of at least tens of millions, the term $N_BTC$ is significantly smaller than $N_BP_\text{param}$. Thus, the overhead introduced by our method is negligible compared to the standard training complexity. Note that the operation in line 5 of \cref{algorithm:method} can be efficiently performed offline before training by using larger batches in inference mode. As empirically detailed in \cref{tab:training_time_overhead}, {\method} incurs a modest computational overhead.

\begin{algorithm}[!t]
  \caption{\method.} \label{algorithm:method}
    {\bf Input:} Target Model $f$, training dataset $\mathcal{D}$, models $f_L$ and $f_N$ trained from $\mathcal{D}$ by learning algorithms $A_L$ and $A_N$, loss weight $\alpha$, Sinkhorn iteration step $T$.
    \begin{algorithmic}[1]
        \STATE Randomly initialize $f$.
        \FOR{epoch $=1,2,\ldots$}
        \FOR{mini-batch $=1,2,\ldots$}
        \STATE Initialize scaling coefficients vectors $u \leftarrow \frac{1}{C}\mathbf{1}_C, v \leftarrow \frac{1}{N_B}\mathbf{1}_{N_B}$. \quad \textcolor{gray}{// $N_B$ is the total number of samples in current mini-batch}
        \STATE Get predictions of $f_L$ and $f_N$ for the mini-batch.  \quad \textcolor{gray}{// Can be precomputed and stored offline for the entire dataset before training}
        \STATE Get predictions of $f$ for the mini-batch.
        \STATE Calculate $M = e^{-\frac{P}{2}}$, $P = [p_1,\ldots,p_{N_B}]$, $p_i = -\log f_N(\bm{x}_i) - \log f(\bm{x}_i)$
        \FOR{$t=1,2,\ldots,T$} 
            \STATE $u \leftarrow \frac{\sum_{i=1}^N f_L(\bm{x}_i)}{N}\ ./ \ (M v)$.
            \STATE $v \leftarrow \frac{\mathbf{1}_N}{N}\ ./ \ (M^T u)$. \quad \textcolor{gray}{// Sinkhorn-Knopp iteration}
        \ENDFOR
        \STATE $Q = N_B\cdot \mathrm{diag}(u) M \mathrm{diag}(v)$. \quad  \textcolor{gray}{// Outer loop optimization for $Q$}
        \STATE Compute $\mathcal{L}_{\mathrm{Base}}$ and $\mathcal{L}_{\mathrm{\method}}$ for the mini-batch.
        \STATE Train $f$ by minimizing $\mathcal{L}_{\mathrm{Overall}} = \mathcal{L}_{\mathrm{Base}} + \alpha \mathcal{L}_{\mathrm{\method}}$. \quad  \textcolor{gray}{// Inner loop optimization for $f$}
        \ENDFOR
        \ENDFOR
        
    \end{algorithmic}
\end{algorithm}

\section{Experiments}

\begin{table}[t]
\centering
\caption{Test accuracy (\%) on CIFAR-10 and CIFAR-100 under symmetric label noise. The table compares various methods across different class imbalance ratios (10, 100) and noise ratios (0.4, 0.6). Best results are highlighted in bold.}
\label{tab: sym-cifar}
\begin{tabular}{llcccccccc}
\toprule
\multicolumn{2}{c}{Noise mode}                          & \multicolumn{8}{c}{Symmetric}                                                                                                               \\ 
\midrule
\multicolumn{2}{c}{Dataset} & \multicolumn{4}{c}{CIFAR-10} & \multicolumn{4}{c}{CIFAR-100} \\ 
\midrule
\multicolumn{2}{c}{Imbalance ratio} & \multicolumn{2}{c}{10} & \multicolumn{2}{c}{100} & \multicolumn{2}{c}{10} & \multicolumn{2}{c}{100} \\ 
 \midrule
\multicolumn{2}{c}{Noise ratio} & 0.4 & 0.6 & 0.4 & 0.6 & 0.4 & 0.6 & 0.4 & 0.6 \\ \midrule
\multicolumn{2}{c}{CE} & 71.67 & 61.16 & 47.81 & 28.04 & 34.53 & 23.63 & 21.99 & 15.51 \\ \midrule
\multicolumn{1}{l|}{\multirow{2}{*}{NLL}} & DivideMix & 80.20 & 79.94 & 32.42 & 34.73 & 47.93 & 41.65 & 36.20 & 26.29 \\
\multicolumn{1}{l|}{}  & UNICON & 80.81 & 77.20 & {\textbf{67.51}} & {\textbf{59.04}} & 52.34 & 44.87 & 34.01 & 27.33 \\ \midrule
\multicolumn{1}{l|}{\multirow{3}{*}{LTL}} & LA & 72.32 & 64.23 & 52.05 & 42.64 & 34.28 & 24.07 & 22.59 & 16.40 \\
\multicolumn{1}{l|}{} & LDAM & 74.59 & 68.17 & 49.79 & 36.46 & 32.29 & 26.60 & 18.81 & 12.65 \\
\multicolumn{1}{l|}{} & IB & 72.66 & 62.19 & 41.93 & 34.17 & 34.85 & 22.81 & 20.34 & 12.10 \\ \midrule
\multicolumn{1}{l|}{\multirow{6}{*}{LTNLL}} & TABASCO & \underline{85.47} & \underline{84.83} & 62.76 & 55.49 & \underline{56.89} & \underline{45.68} & \underline{36.92} & \underline{28.50} \\
 \multicolumn{1}{l|}{} & MW-Net & 70.34 & 58.48 & 45.54 & 40.03 & 32.29 & 21.71 & 20.76 & 14.27 \\
 \multicolumn{1}{l|}{} & FR & 70.19 & 60.86 & 49.36 & 30.15 & 30.24 & 16.99 & 22.56 & 15.07 \\
 \multicolumn{1}{l|}{} & HAR & 74.11 & 60.92 & 51.61 & 37.96 & 36.53 & 24.68 & 20.41 & 15.03 \\
 \multicolumn{1}{l|}{} & RoLT & 81.75 & 79.67 & 60.21 & 44.36 & 43.08 & 32.57 & 24.11 & 16.59 \\ 
 \multicolumn{1}{l|}{} & RCAL &83.43&79.89 & 61.78 & 55.31& 57.20 & 43.36  & 33.36 & 20.08\\
 \midrule
\multicolumn{2}{c}{\multirow{2}{*}{D-SINK (Ours)}} & \textbf{89.00} & \textbf{86.93} & {\underline{66.43}} & {\underline{57.80}} & \textbf{58.96} & \textbf{48.26} & \textbf{38.66} & \textbf{30.42} \\ 
&  & $\pm$0.28 & $\pm$0.33 & $\pm$0.37 & $\pm$0.45 & $\pm$0.31 & $\pm$0.41 & $\pm$0.27 & $\pm$0.31\\
\bottomrule
\end{tabular}
\end{table}

\begin{table}[t]
\centering
\caption{Test accuracy (\%) on CIFAR-10 and CIFAR-100 under asymmetric noise. The table compares various methods across different class imbalance ratios (10, 100) and noise ratios (0.2, 0.4). Best results are highlighted in bold.}
\label{tab: asym-cifar}
\begin{tabular}{llcccccccc}
\toprule
\multicolumn{2}{c}{Noise mode}                          & \multicolumn{8}{c}{Asymmetric}                                                                                                               \\ \midrule
\multicolumn{2}{c}{Dataset} & \multicolumn{4}{c}{CIFAR-10} & \multicolumn{4}{c}{CIFAR-100} \\ \midrule
\multicolumn{2}{c}{Imbalance ratio} & \multicolumn{2}{c}{10} & \multicolumn{2}{c}{100} & \multicolumn{2}{c}{10} & \multicolumn{2}{c}{100} \\ \midrule
\multicolumn{2}{c}{Noise ratio}                         & 0.2            & 0.4            & 0.2            & 0.4            & 0.2            & 0.4            & 0.2            & 0.4            \\ \midrule
\multicolumn{1}{l|}{Baseline} & CE & 79.90 & 62.88 & 56.56 & 44.64 & 44.45 & 32.05 & 25.35 & 17.89 \\ \midrule
\multicolumn{1}{l|}{\multirow{2}{*}{NLL}} & DivideMix & 80.12 & 74.73 & 41.12 & 42.79 & 54.83 & 41.95 & 36.46 & 29.69 \\
\multicolumn{1}{l|}{} & UNICON & 75.01 & 72.89 & 44.48 & 32.64 & 57.19 & 45.43 & 37.78 & 30.56 \\ \midrule
\multicolumn{1}{l|}{\multirow{3}{*}{LT}} & LA & \underline{84.92} & 76.17 & 58.26 & 50.30 & 49.78 & 36.24 & 29.07 & 27.58 \\
 \multicolumn{1}{l|}{} & LDAM & 78.74 & 68.68 & 56.29 & 38.83 & 44.43 & 31.02 & 23.22 & 21.04 \\
 \multicolumn{1}{l|}{} & IB & 70.88 & 60.32 & 50.04 & 31.37 & 47.48 & 33.81 & 27.85 & 20.05 \\ \midrule
\multicolumn{1}{l|}{\multirow{6}{*}{LTNLL}} & TABASCO & 82.13 & \underline{80.57} & \underline{60.35} & \underline{51.19} & \underline{59.45} & \underline{50.43} & \underline{38.43} & \underline{32.15} \\
\multicolumn{1}{l|}{} & MW-Net & 78.46 & 64.82 & 59.37 & 45.21 & 43.36 & 31.25 & 27.56 & 20.04 \\
 \multicolumn{1}{l|}{}& FR & 82.23 & 69.13 & 56.62 & 45.12 & 47.01 & 36.09 & 25.27 & 20.46 \\
 \multicolumn{1}{l|}{}& HAR & 78.75 & 70.23 & 58.89 & 49.97 & 49.36 & 35.98 & 27.90 & 20.03 \\
 \multicolumn{1}{l|}{}& RoLT & 80.38 & 78.39 & 54.79 & 50.41 & 50.62 & 39.32 & 33.16 & 25.73 \\ 
\multicolumn{1}{l|}{} & RCAL & 81.65 & 78.78 & 56.60 & 49.85 & 56.29 & 42.86 & 36.15 & 26.90 \\ 
\midrule
\multicolumn{2}{c}{\multirow{2}{*}{D-SINK (Ours)}} & \textbf{87.09} & \textbf{84.36} & \textbf{61.98} & \textbf{53.06} & \textbf{61.69} & \textbf{51.56} & \textbf{40.10} & \textbf{34.06} \\
 & & $\pm$0.32 & $\pm$0.29 & $\pm$0.24 & $\pm$0.46 & $\pm$0.31 & $\pm$0.41 & $\pm$0.34 & $\pm$0.38 \\
\bottomrule
\end{tabular}
\end{table}

\subsection{Experimental Setup}
\textbf{Datasets.} To evaluate the robustness of our method on long-tailed noisy datasets, we conducted experiments on several datasets, including those with synthetic noise and real-world noise in long-tailed distributions. Following \citet{DBLP:journals/corr/abs-2108-11569,DBLP:conf/iccv/LuZHCW23}, we constructed long-tailed datasets with synthetic noise based on the commonly used \textit{CIFAR-10} and \textit{CIFAR-100}~\citep{krizhevsky2009learning}. Specifically, following \citet{DBLP:conf/cvpr/CuiJLSB19}, we first construct long-tailed classes using exponential sampling with varying imbalance ratios. Next, we inject synthetic noise into the long-tailed versions of the clean datasets, including commonly used symmetric and asymmetric noise~\citep{DBLP:conf/nips/ShuXY0ZXM19} with varying noise ratios. We conduct extensive experiments under different settings, combining varying noise types (symmetric and asymmetric), imbalance ratios, and noise ratios. The definitions of imbalance ratio ($\mathrm{IR}$) and noise ratio ($\mathrm{NR}$) can be found in \cref{subsec: Problem Formulation}. We also evaluate the effectiveness of {\method} on four datasets with real-world noisy labels: \textit{CIFAR-10N}~\citep{DBLP:conf/iclr/WeiZ0L0022}, \textit{CIFAR-100N}~\citep{DBLP:conf/iclr/WeiZ0L0022}, \textit{Red Mini-ImageNet}~\citep{DBLP:conf/icml/JiangHLY20}, and \textit{Clothing1M}~\citep{DBLP:conf/cvpr/XiaoXYHW15}. We use exponential sampling to construct their long-tailed versions with imbalance ratio of 10.

\textbf{Baselines.} We evaluate our method against a comprehensive set of baselines: (1) the vanilla baseline: empirical risk minimization with cross-entropy loss (CE); (2) noisy-label learning methods including DivideMix~\citep{DBLP:conf/iclr/LiSH20} and UNICON~\citep{DBLP:conf/cvpr/KarimRRMS22}; (3) long-tailed learning methods including LA~\citep{DBLP:conf/iclr/MenonJRJVK21}, IB~\citep{DBLP:conf/iccv/ParkLJ021}, and LDAM~\citep{DBLP:conf/nips/CaoWGAM19}; (4) long-tailed noisy-label learning methods including TABASCO~\citep{DBLP:conf/iccv/LuZHCW23}, MW-Net~\citep{DBLP:conf/nips/ShuXY0ZXM19}, FR~\citep{DBLP:journals/corr/abs-2303-12291}, HAR~\citep{DBLP:conf/iclr/CaoCLAGM21}, RoLT~\citep{DBLP:journals/corr/abs-2108-11569} and RCAL~\citep{DBLP:conf/iccv/ZhangZYY023}. This extensive comparison encompasses a broad spectrum of state-of-the-art methods in long-tailed noisy-label learning and related domains.

\noindent\textbf{Implementation Details.} We use an 18-layer ResNet as the backbone. Standard data augmentations are applied.
The mini-batch size is set to 64, and all methods are trained using SGD with a momentum of 0.9 and a weight decay of 5e-4. For all methods, the models are trained for 300 epochs to ensure fair comparison. The initial learning rate is set to 0.02 and decays by a factor of 10 every 50 epochs after the first 100 epochs. For {\method}, unless otherwise specified, we use the objective from \citet{DBLP:conf/iccv/LuZHCW23} as $\mathcal{L}_{\mathrm{base}}$  in \cref{Eq: objective} and leverage our {\method} framework to further improve the robustness against noisy labels and long-tailed distributions. 
We set $\alpha = 10^{-3}$ in \cref{Eq: objective}, to ensure that the two losses,  $\mathcal{L}_{\mathrm{base}}$  and  $\mathcal{L}_{\mathrm{\method}}$, are on roughly the same scale. We choose IB and DivideMix as $A_L$ and $A_N$ to get $f_L$ and $f_N$.

\subsection{Main Results}
\textbf{CIFAR-10 and CIFAR-100 (synthetic noise).} We evaluate the performance of different methods on long-tailed noisy-label CIFAR-10 and CIFAR-100 datasets under synthetic noise (symmetric in \cref{tab: sym-cifar}  and asymmetric in \cref{tab: asym-cifar}), with varying imbalance ratios and noise ratios. For the 11 baseline methods we considered, we observe 
that: (1) Except for the noisy-label learning method UNICON and the long-tailed noisy-label learning methods TABASCO and RoLT, all other baselines fail to consistently outperform CE across all settings.  (2) No method can maintain the best performance across all settings among the 11 baselines. While the TABASCO performs well in most scenarios compared to other methods, it significantly underperforms LA on CIFAR-10 with asymmetric noise, imbalance ratio 10, and noise ratio 0.2. 
These highlight the difficulty of the long-tailed noisy-label learning scenario, and also demonstrate that existing methods, whether attempting to address long-tailed distributions or noisy labels individually, or combining solutions for both, fail to effectively tackle the deep coupling of these two challenges, resulting in suboptimal performance.

For {\method}, we observe:
(1) {\method} achieves the best results across all 16 different settings, with significant improvements over the second-best results in each setting.
(2) Unlike traditional knowledge distillation paradigms~\citep{DBLP:journals/ijon/LiuZW20,DBLP:conf/iclr/TanRHQZL19}, where the performance of the student model is enhanced by distilling from \emph{strong} teachers, the two ``teacher" models ($f_L$ and $f_N$) in our method perform comparably to CE and, in some settings, even worse. This indicates that our Dual-granularity Distillation framework effectively extracts and integrates knowledge from weak but specialized teacher models.

\begin{table}[t]
\centering
\caption{Test accuracy (\%) comparison on real-world noisy datasets (CIFAR-10N, CIFAR-100N, Red Mini-ImageNet, and Clothing1M). The best results are highlighted in bold.}
\label{tab: real world-main}
\begin{tabular}{llcccccc}
\toprule
\multicolumn{6}{c}{Real-word noisy annotations}\\ \midrule
\multicolumn{2}{c}{Dataset}          & \multicolumn{2}{c}{CIFAR-10N}   & \multicolumn{2}{c}{CIFAR-100N}  & Red Mini-ImageNet      & Clothing1M   \\ \midrule
\multicolumn{2}{c}{Imbalance ratio} & 10 & 100 & 10 & 100 & \multicolumn{2}{c}{10}                                    \\ 
 \midrule
\multicolumn{2}{c}{Noise ratio}      & \multicolumn{4}{c}{$\approx 0.4$}   &  0.4           & $\approx 0.4$            \\ \midrule
\multicolumn{1}{c|}{Baseline}        & CE         & 63.44   &49.86    & 38.1  &29.26      & 31.46         & 46.41          \\ \midrule
\multicolumn{1}{l|}{\multirow{2}{*}{NLL}} & DivideMix  & 70.35  & 55.48     & 46.87  & 38.73      & 41.64         & 71.09         \\
\multicolumn{1}{l|}{}                & UNICON     & 75.62    & \underline{62.75}   & \underline{55.32}  & 40.02     & 38.48         & 72.34          \\ \midrule
\multicolumn{1}{l|}{\multirow{3}{*}{LT}} & LA         & 70.47   & 60.69    & 38.67   & 30.83    & 35.52         & 53.28         \\
\multicolumn{1}{l|}{}                & LDAM       & 61.94   & 51.68    & 38.37  & 30.59     & 32.48         & 54.01         \\
\multicolumn{1}{l|}{}                & IB         & 62.56   & 52.06    & 42.34  & 32.69     & 34.56         & 52.46          \\ \midrule
\multicolumn{1}{l|}{\multirow{6}{*}{LTNLL}} & TABASCO    & \underline{80.73}   & 62.47    & 53.67   & \underline{40.70}    & \underline{49.68}         & \underline{75.26}        \\
\multicolumn{1}{l|}{}                & MW-Net     & 69.73   & 53.69    & 44.67   & 34.40    & 40.26         & 58.74         \\
\multicolumn{1}{l|}{}                & FR         & 67.15   & 49.02    & 39.44   & 30.37    & 33.27         & 47.04       \\
\multicolumn{1}{l|}{}                & HAR        & 75.87   & 59.18    & 44.69   & 33.96    & 38.71         & 55.28       \\
\multicolumn{1}{l|}{}                & RoLT       & 74.84   & 58.37    & 47.02  & 36.21     & 47.37         & 63.79          \\ 
\multicolumn{1}{l|}{}                & RCAL       & 76.74   & 60.62    & 48.35  & 37.23     & 46.40         & 69.33         \\ 
\midrule
\multicolumn{2}{c}{\multirow{2}{*}{D-SINK (Ours)}}   & \textbf{82.89} & \textbf{66.12} & \textbf{55.43} & \textbf{42.51} & \textbf{51.46} & \textbf{76.94} \\ 
& &$\pm$0.28 & $\pm$0.33 & $\pm$0.27 & $\pm$0.35 & $\pm$0.26 & $\pm$0.32 \\
\bottomrule
\end{tabular}
\end{table}

\noindent\textbf{CIFAR-10N/100N, Red Mini-ImageNet, and Clothing1M (real-world noise).}  In \cref{tab: real world-main}, we present the performance of different methods on long-tailed noisy datasets with real-world noise, including CIFAR-10N, CIFAR-100N, Red Mini-ImageNet, and Clothing1M. 
The complex and often instance-dependent nature of the noise in these real-world datasets typically poses a more substantial challenge to model robustness compared to synthetic noise configurations. Encouragingly, the results demonstrate that even in these more demanding and realistic scenarios, {\method} consistently outperforms competing methods across all four datasets. This consistent strong performance highlights that {\method} is exceptionally good at dealing with real-world label noise along with class imbalances, proving its practical value.

\begin{table}[t]
\caption{Mutiple metrics  on CIFAR-100 (imbalance ratio 10) under different symmetric noise (ratio 0.4 and 0.6). Performance is detailed for Many, Medium, and Few splits, along with the overall accuracy(\%). We also report Macro-F1 and AUC. Our method (D-SINK) demonstrates leading performance, particularly in Medium, Few, overall accuracies, Macro-F1 and AUC.}
\label{tab: per-split-results}
\centering

\begin{tabular}{ccccccc}
\toprule
\multicolumn{1}{c}{Noise ratio (Sym)} & \multicolumn{6}{c}{0.4}  \\ 
\midrule
\multicolumn{1}{c}{Split} & Many Acc & Medium Acc & Few Acc & Overall Acc & Macro-F1 & AUC \\
\midrule
\multicolumn{1}{c}{CE} & 46.77 & 31.43 & 16.93 & 34.53 & 0.29 & 0.85 \\
\multicolumn{1}{c}{DivideMix} & \textbf{74.02} & 55.58 & 9.52 & 47.93 & 0.46 & 0.94 \\
\multicolumn{1}{c}{IB} & 42.77 & 34.77 & 26.48 & 34.85 & 0.34 & 0.88 \\
\multicolumn{1}{c}{TABASCO} & 66.03 & 55.12 & 41.24 & 56.89 & 0.53 & 0.95 \\
\multicolumn{1}{c}{D-SINK(ours)} & 68.03 & \textbf{57.82} & \textbf{44.11} & \textbf{58.96} & \textbf{0.56} & \textbf{0.97} \\
\midrule
\multicolumn{1}{c}{Noise ratio (Sym)} & \multicolumn{6}{c}{0.6}  \\ 
\midrule
\multicolumn{1}{c}{Split} & Many Acc & Medium Acc & Few Acc & Overall Acc & Macro-F1 & AUC \\
\midrule
\multicolumn{1}{c}{CE} & 35.29 & 17.21 & 6.93 & 23.63 & 0.21 & 0.79 \\
\multicolumn{1}{c}{DivideMix} & \textbf{65.94} & 45.95 & 9.76 & 41.65 & 0.44 & 0.93 \\
\multicolumn{1}{c}{IB} & 27.90 & 22.57 & 17.69 & 22.81 & 0.23 & 0.82 \\
\multicolumn{1}{c}{TABASCO} & 57.26 & 46.65 & 35.52 & 45.68 & 0.45 & 0.94 \\
\multicolumn{1}{c}{D-SINK(ours)} & 59.45 & \textbf{47.98} & \textbf{36.59} & \textbf{48.26} & \textbf{0.49} & \textbf{0.96} \\
\bottomrule
\end{tabular}
\end{table}

\subsection{Further Analysis}
\textbf{Many/Medium/Few Per-split Analysis.} To specifically analyze the robustness of different methods to long-tailed distributions in the presence of noisy labels, \cref{tab: per-split-results} presents more fine-grained metrics on many, medium, and few class splits. Specifically, we conduct experiments on CIFAR-100 with symmetric noise, an imbalance ratio of 10, and noise ratios of 0.4 and 0.6. Many, medium, and few refer to the splits of CIFAR-100’s 100 classes: the 30 classes with the highest number of samples, the 40 classes with a moderate number of samples, and the 30 classes with the fewest samples, respectively. Alongside CE, we include three baselines in \cref{tab: per-split-results}: DivideMix, IB, and TABASCO, representing noisy-label learning, long-tailed learning, and long-tailed noisy-label learning methods, respectively. It can be observed that noisy-label learning methods, by emphasizing the discarding of hard samples, end up discarding too many tail samples, exacerbating the imbalance performance. On the other hand, long-tailed learning methods focus on balancing predictions between head and tail classes but do not distinguish noisy samples, leading to a sacrifice in the performance of head classes. The state-of-the-art long-tailed noisy-label learning method, TABASCO, achieves good and balanced performance compared to other baselines. {\method} further improves upon TABASCO, consistently showing better performance across each split and various evaluation metrics.

\begin{table}[!t]
\centering
\caption{Test accuracy (\%) demonstrating the performance improvement achieved by integrating {\method} with various baselines (CE, HAR, RoLT, TABASCO). Experiments are conducted on CIFAR-10/100 under symmetric noise, with imbalance ratios of 10 and 100, and noise ratios of 0.4 and 0.6.}

\label{tab: sym-improve}
\begin{tabular}{llcccccccc}
\toprule
Noise mode       & \multicolumn{8}{c}{Symmetric}                                                                                                            \\ \midrule
\multicolumn{1}{l}{Dataset} & \multicolumn{4}{c}{CIFAR-10} & \multicolumn{4}{c}{CIFAR-100} \\ \midrule
\multicolumn{1}{l}{Imbalance ratio} & \multicolumn{2}{c}{10} & \multicolumn{2}{c}{100} & \multicolumn{2}{c}{10} & \multicolumn{2}{c}{100} \\ \midrule
Noise ratio      & 0.4            & 0.6            & 0.4            & 0.6            & 0.4            & 0.6            & 0.4            & 0.6            \\ \midrule
CE & 71.67 & 61.16 & 47.81 & 28.04 & 34.53 & 23.63 & 21.99 & 15.51 \\
\quad + {\method} & \textbf{76.09} & \textbf{64.17} & \textbf{53.85} & \textbf{41.83} & \textbf{42.74} & \textbf{29.13} & \textbf{24.69} & \textbf{17.23} \\ \midrule
HAR & 74.11 & 60.92 & 51.61 & 37.96 & 36.53 & 24.68 & 20.41 & 15.03 \\
\quad + {\method} & \textbf{76.32} & \textbf{65.03} & \textbf{53.20} & \textbf{39.49} & \textbf{38.43} & \textbf{26.12} & \textbf{22.05} & \textbf{16.52} \\ \midrule
RoLT & 81.75 & 79.67 & 60.21 & 44.36 & 43.08 & 32.57 & 24.11 & 16.59 \\
\quad + {\method} & \textbf{87.20} & \textbf{85.99} & \textbf{64.29} & \textbf{50.87} & \textbf{51.24} & \textbf{38.57} & \textbf{29.86} & \textbf{22.11} \\ \midrule
TABASCO & 85.47 & 84.83 & 62.76 & 55.49 & 56.89 & 45.68 & 36.92 & 28.50 \\
\quad + {\method} & \textbf{89.00} & \textbf{86.93} & \textbf{66.43} & \textbf{57.80} & \textbf{58.96} & \textbf{48.26} & \textbf{38.66} & \textbf{30.42} \\ \bottomrule
\end{tabular}
\end{table}

\begin{table}[t]
\centering
\caption{Test accuracy (\%) demonstrating the performance improvement achieved by integrating {\method} with various baselines (CE, HAR, RoLT, TABASCO). Experiments are conducted on CIFAR-10/100 under asymmetric noise, with imbalance ratios of 10 and 100, and noise ratios of 0.4 and 0.6.}

\label{tab: asym-improve}
\begin{tabular}{llcccccccc}
\toprule
Noise mode       & \multicolumn{8}{c}{Asymmetric}                                  \\ \midrule
\multicolumn{1}{l}{Dataset} & \multicolumn{4}{c}{CIFAR-10} & \multicolumn{4}{c}{CIFAR-100} \\ \midrule
\multicolumn{1}{l}{Imbalance ratio} & \multicolumn{2}{c}{10} & \multicolumn{2}{c}{100} & \multicolumn{2}{c}{10} & \multicolumn{2}{c}{100} \\ \midrule
Noise ratio      & 0.2           & 0.4          & 0.2           & 0.4           & 0.2           & 0.4          & 0.2           & 0.4           \\ \midrule
CE & 79.9 & 62.88 & 56.56 & 44.64 & 44.45 & 32.05 & 25.35 & 17.89 \\
\quad + {\method} & \textbf{82.46} & \textbf{65.56} & \textbf{58.15} & \textbf{47.43} & \textbf{52.34} & \textbf{38.06} & \textbf{27.03} & \textbf{19.68} \\ \midrule
HAR & 78.75 & 70.23 & 58.89 & 49.97 & 49.36 & 35.98 & 27.9 & 20.03 \\
\quad + {\method} & \textbf{80.72} & \textbf{72.11} & \textbf{60.49} & \textbf{51.62} & \textbf{51.04} & \textbf{37.58} & \textbf{29.44} & \textbf{21.55} \\ \midrule
RoLT & 80.38 & 78.39 & 54.79 & 50.41 & 50.62 & 39.32 & 33.16 & 25.73 \\
\quad + {\method} & \textbf{86.09} & \textbf{83.78} & \textbf{60.40} & \textbf{54.83} & \textbf{59.64} & \textbf{42.87} & \textbf{38.42} & \textbf{29.15} \\ \midrule
TABASCO & 82.13 & 80.57 & 60.35 & 51.19 & 59.45 & 50.43 & 38.43 & 32.15 \\
\quad + {\method} & \textbf{87.09} & \textbf{84.36} & \textbf{61.98} & \textbf{53.06} & \textbf{61.69} & \textbf{51.56} & \textbf{40.10} & \textbf{34.06} \\ \bottomrule
\end{tabular}
\end{table}

\noindent\textbf{{\method} as A Universal Framework.} \cref{tab: sym-improve} and \cref{tab: asym-improve} summarizes the performance of our {\method} when integrated with different baseline methods (using the objective of the method for $\mathcal{L}_{\mathrm{Base}}$ in \cref{Eq: objective}), including CE, HAR, RoLT, and TABASCO. It is clear that our {\method} can significantly improve the long-tailed noisy-label classification performance of different baselines. Furthermore, due to the orthogonal perspective of our {\method} of other methods, it can consistently benefit from improved potential future methods.

\begin{figure}[!t]
\centering
\subfigure[Ablation on $\alpha$.]{
\includegraphics[height=0.2\textwidth]{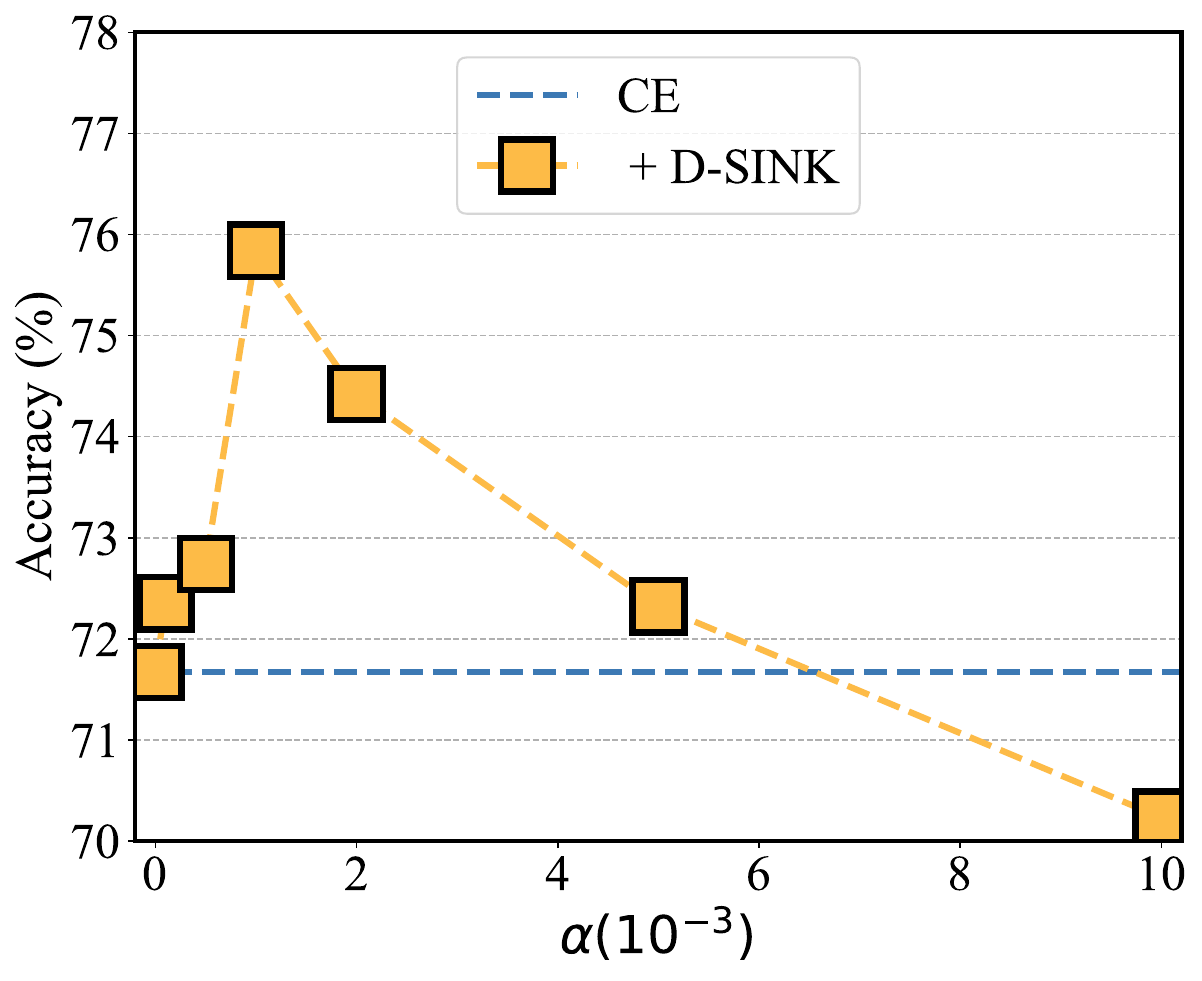}
\label{fig:ablation_alpha_results}
}
\subfigure[CIFAR-10 with symmetric noise]{
\includegraphics[height=0.2\textwidth]{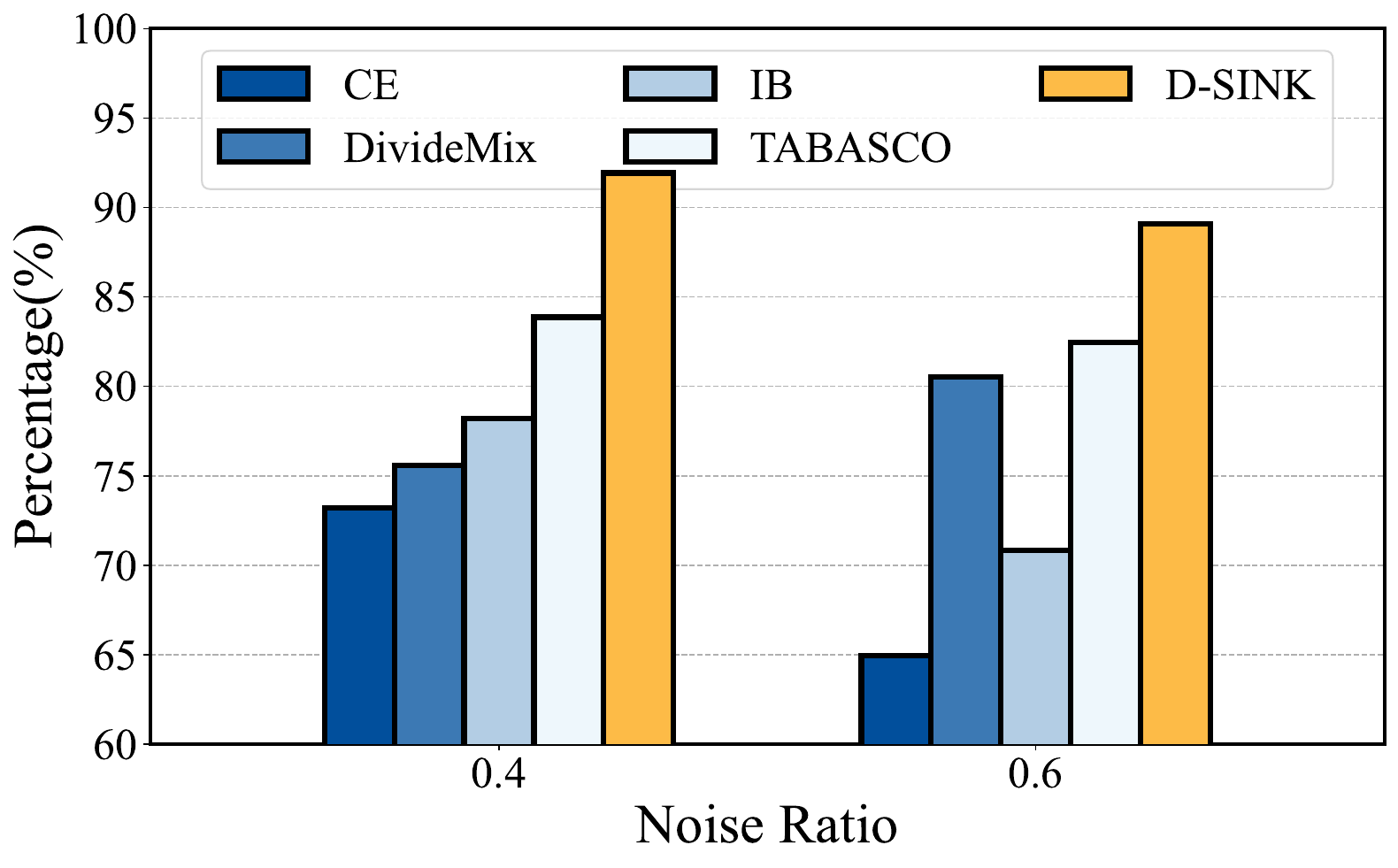}
\label{fig: Noise identify}
}
\subfigure[CIFAR-10 with asym noise]{
\includegraphics[height=0.2\textwidth]{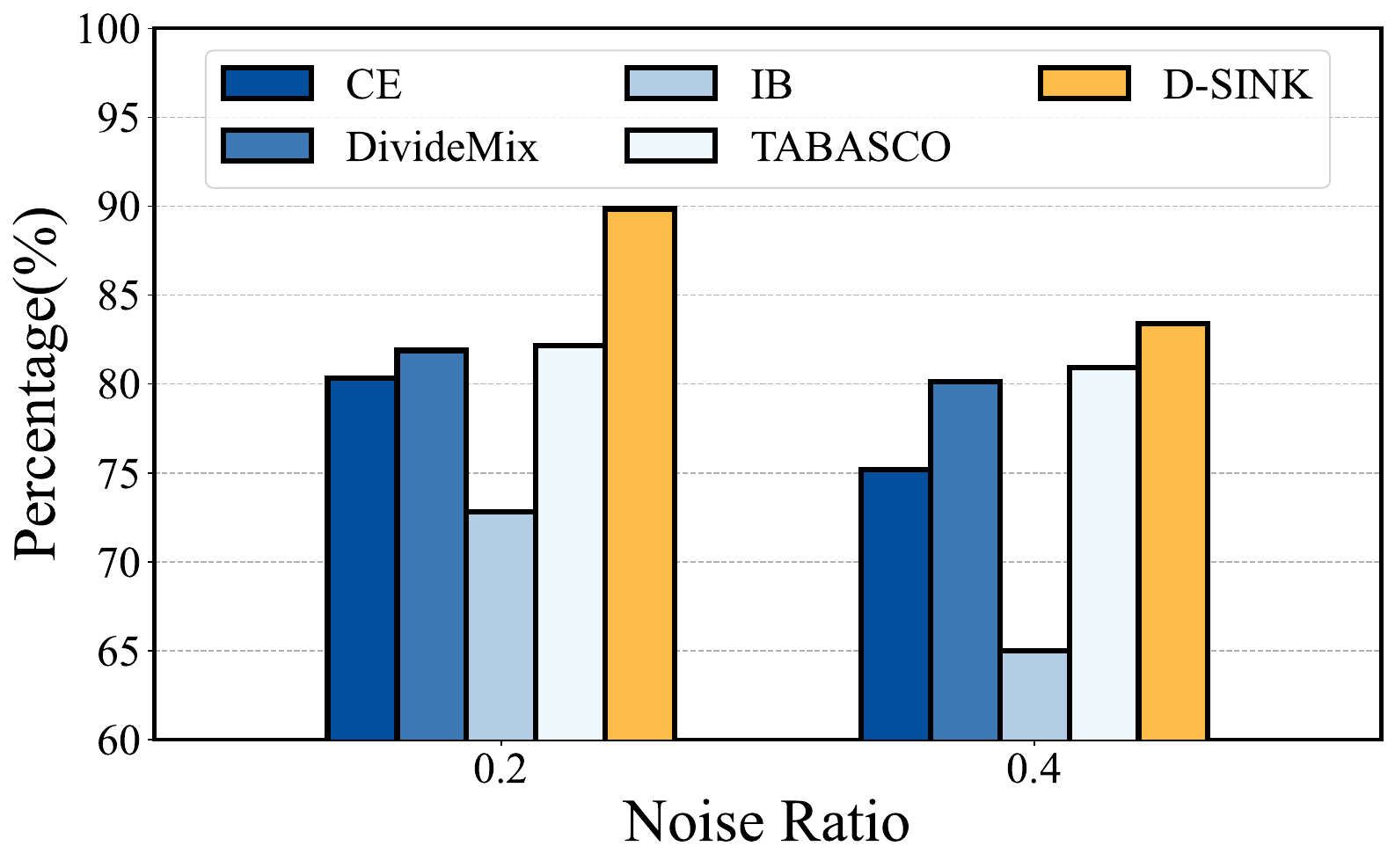}
\label{fig: Noise identify2}
}
\caption{(a) Ablation on hyperparameter $\alpha$. (b) Percentage of noisy training labels corrected (i.e., true labels predicted by the model) on CIFAR-10 with the imbalance ratio of  10 and symmetric noise. (c) Percentage of noisy training labels corrected (i.e., true labels predicted by the model) on CIFAR-10 with the imbalance ratio of  10 and asymmetric noise.}
\end{figure}

\noindent\textbf{Comparison with Naive Distillation and Direct Ensemble.} To demonstrate {\method}'s effectiveness in harmonizing and synergistically utilizing auxiliary models, we compare it against two intuitive baselines: (i) Naive Distillation, which employs the standard distillation loss $\mathcal{L}_{\text{Dist}}$ (\cref{eq: naive distillation}) in place of our proposed {\method} loss in \cref{Eq: objective}; and (ii) Direct Ensemble, which averages the outputs of the two auxiliary models. It is pertinent to note that Direct Ensemble necessitates both auxiliary models during deployment, thereby incurring additional overhead. As illustrated in \cref{tab:ablation study}, {\method} significantly outperforms both Naive Distillation and Direct Ensemble, demonstrating its success in harmonizing individually robust models and amalgamating their insights to improve the target model.

\noindent\textbf{Noisy Sample Prediction Analysis.} To analyze the robustness of different methods to label noise under long-tailed distributions, \cref{fig: Noise identify,fig: Noise identify2} shows the accuracy of predictions on noisy training samples. This highlights the error-correction ability of each method when handling noisy labels during training. Specifically, we conduct experiments on CIFAR-10 (imbalance ratio $\mathrm{IR} = 10$) with symmetric and asymmetric noise at different noise ratios. It can be observed that {\method} significantly outperforms other baselines in all settings, demonstrating superior robustness against noisy labels.

\noindent\textbf{Ablation on Hyperparameter $\alpha$ in \cref{Eq: objective}.} We conduct an ablation study to evaluate the robustness of {\method} to the hyperparameter $\alpha$ defined in \cref{Eq: objective}. This parameter balances the contributions of $\mathcal{L}_{\mathrm{Base}}$ and $\mathcal{L}_{\mathrm{\method}}$. $\alpha$ is systematically varied in $\{0,0.1,0.2,0.5,1,2\} \times 10^{-3}$, incorporating a $10^{-3}$ scaling factor to ensure $\mathcal{L}_{\mathrm{Base}}$ and $\mathcal{L}_{\mathrm{\method}}$ maintained a comparable order of magnitude. The results, presented in \cref{fig:ablation_alpha_results}, demonstrate that {\method} is remarkably robust to variations in $\alpha$: performance remained consistently high and stable across a wide range of $\alpha$. Such insensitivity to $\alpha$ highlights the stability of {\method}, thereby simplifying its practical deployment by alleviating the need for extensive hyperparameter tuning.

\noindent\textbf{Ablation study of $\mathcal{L}_\text{Base}$ and $f_L$.} In \cref{tab:ablation study}, we conduct an ablation study on $\mathcal{L}_\text{Base}$ and $f_L$ to demonstrate their respective contributions. Specifically, $\mathcal{L}_\text{Base}$ equips the model with its fundamental classification capability, while $f_L$ is crucial for addressing the class imbalance inherent in long-tailed distributions. As shown, the performance of our method sees a significant degradation in the absence of either component, confirming their individual importance.

\begin{table}[!t]
\centering
\caption{Ablation study of different Auxiliary Model selections for {\method} with six combinations of an $f_L$ component (LA, IB, or LDAM) and an $f_N$ component, where -D signifies $f_N$ using DivideMix and -U signifies $f_N$ using UNICON. Results are presented on CIFAR-100 (imbalance ratio 10) with symmetric noise.}
\label{tab:aux_ablation_results}
\begin{tabular}{c|cc|cccccc}
\toprule
                         &                      &                           & \multicolumn{6}{c}{{\method} (Ours)}                                                                                   \\
\multirow{-2}{*}{Noise Ratio} & \multirow{-2}{*}{CE} & \multirow{-2}{*}{TABASCO} & LA-D & LA-U      & IB-D               & IB-U     & LDAM-D & LDAM-U \\\midrule
0.4                   & 34.53                & 56.89                     & 57.82        & 58.32          & {\color[HTML]{1F2329} 58.86} & \textbf{58.96} & 57.88          & 58.03       \\
0.6                   & 23.63                & 45.68                     & 49.14        & \textbf{49.57} & 48.26                        & 48.56          & 48.35          & 48.79      \\\bottomrule
\end{tabular}
\end{table}

\noindent\textbf{Versatility in Auxiliary Model Selection.}
We evaluate {\method}'s versatility and robustness by pairing it with various auxiliary models. Specifically, we tested three distinct $f_L $models targeting long-tailed recognition (LA, IB, and LDAM), in combination with two $f_N$ models addressing noisy labels (DivideMix and Unicon). Across all six $f_L$-$f_N$	configurations, as detailed in \cref{tab:aux_ablation_results},  {\method} consistently exhibited strong performance and achieved significant improvements. These results underscore {\method}'s robustness to the specific choice of auxiliary component and highlight its flexible integration capabilities, yielding substantial gains regardless of the particular $f_L$  or $f_N$ strategy employed.

\begin{figure}[!t]
\centering
\subfigure[CE]{
\includegraphics[width=0.3\linewidth]{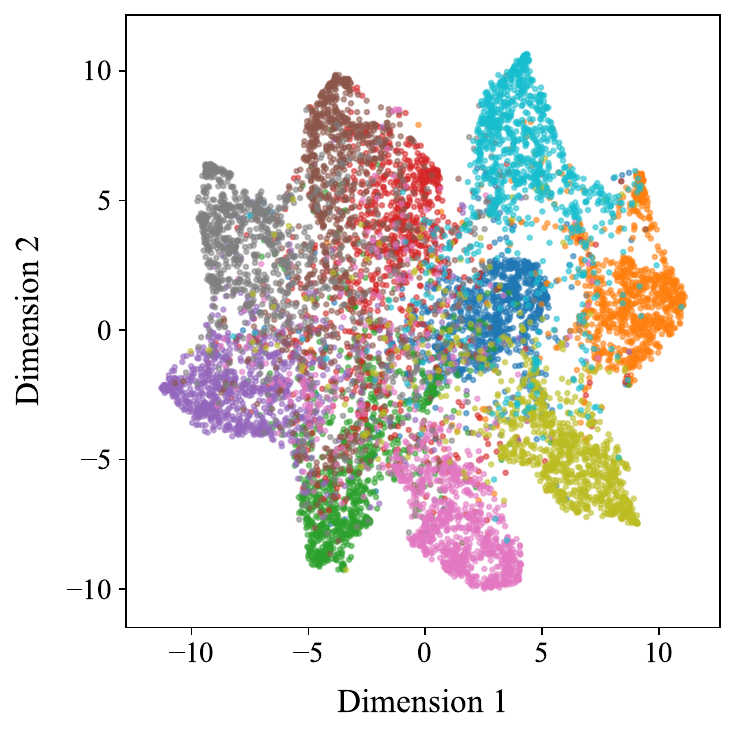}
\label{fig:cliptsne}
}
\subfigure[TABASCO]{
\includegraphics[width=0.3\linewidth]{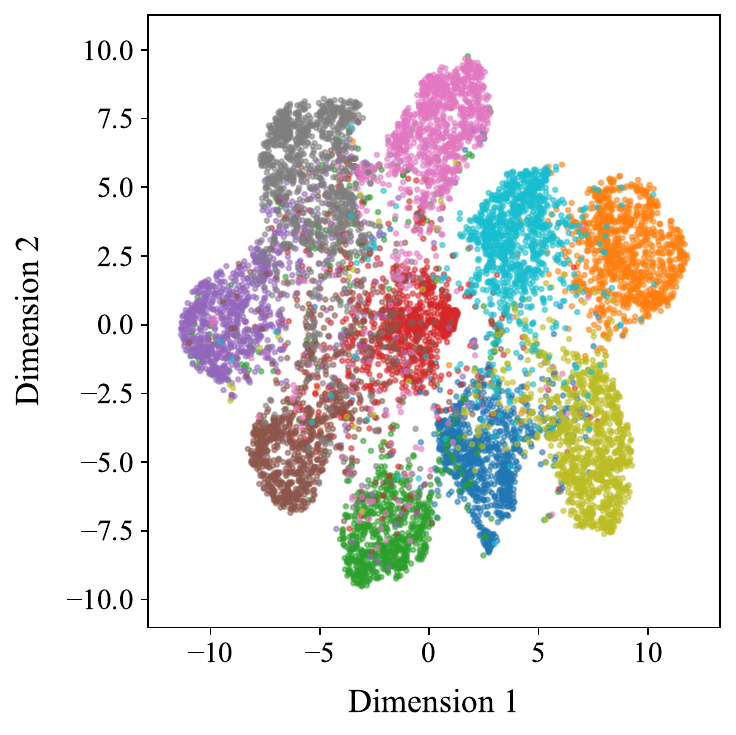}
\label{fig:noetftsne}
}
\subfigure[{\method}]{
\includegraphics[width=0.3\linewidth]{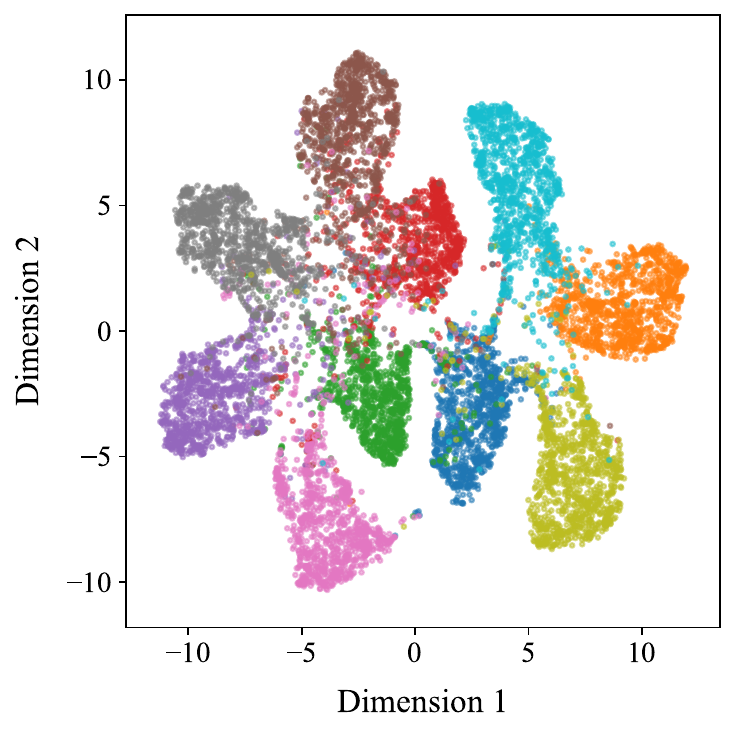}
\label{fig:ourstsne}
}
\caption{T-SNE visualization  of the features learned by CE, TABASCO and
{\method} on CIFAR-10 with imbalance ratio of 0.1 and noise ratio of 0.4. }\label{fig: tsne}
\end{figure}

\noindent\textbf{Representation Quality Analysis.} For a comparative analysis of feature quality, we conduct t-SNE visualizations of the representations learned by CE, TABASCO, and {\method}, presented in \cref{fig: tsne}. These visualizations are based on the CIFAR-10 dataset (with an imbalance ratio of 10 and a symmetric label noise ratio of 0.4). The results indicate that the baseline CE method learns features that poorly separate different classes, demonstrating the distinct difficulties arising from co-occurring class imbalance and label noise. TABASCO, a prominent method specifically designed for long-tailed noisy datasets, moderately improves upon CE's feature quality. In contrast, our {\method}, through its effective dual-granularity robust distillation, achieves a more substantial improvement in the quality of the extracted network features.

\begin{table}[h!]
\centering

\caption{Ablation study of $\mathcal{L}_{\mathrm{Base}}$ and $f_L$ for D-SINK. Results are presented on CIFAR-10 and CIFAR-100 with an imbalance factor of 10. Results of Naive Distillation and Direct Ensemble are also presented. The subsequent ``Margin'' rows quantify the absolute performance improvement of our D-SINK over various baselines.}
\label{tab:ablation study}
\begin{tabular}{llcccccccc}
\toprule
\multicolumn{2}{c}{Noise mode} & \multicolumn{4}{c}{Symmetric} & \multicolumn{4}{c}{Asymmetric} \\
\midrule
\multicolumn{2}{c}{Dataset} & \multicolumn{2}{c}{CIFAR-10} & \multicolumn{2}{c}{CIFAR-100} & \multicolumn{2}{c}{CIFAR-10} & \multicolumn{2}{c}{CIFAR-100} \\
\midrule
\multicolumn{2}{c}{Noise ratio} & 0.4 & 0.6 & 0.4 & 0.6 & 0.2 & 0.4 & 0.2 & 0.4 \\
\midrule
\multicolumn{2}{c}{CE}  & 71.67 & 61.16 & 34.53 & 23.63 & 79.90 & 62.88 & 44.45 & 32.05 \\
\multicolumn{2}{c}{DivideMix ($f_N$)} & 80.20 & 79.94 & 47.93 & 41.65 & 80.12 & 74.73 & 54.83 & 41.95 \\
\multicolumn{2}{c}{IB ($f_L$)} & 72.66 & 62.19 & 34.85 & 22.81 & 70.88 & 60.32 & 47.48 & 33.81 \\
\multicolumn{2}{c}{TABASCO} & 85.47 & 84.83 & 56.89 & 45.68 & 82.13 & 80.57 & 59.45 & 50.43 \\
\midrule
\multicolumn{2}{c}{w/o $\mathcal{L}_{\mathrm{Base}}$} & 10.00 & 10.02 & 4.25 & 5.01 & 10.13 & 10.05 & 5.12 & 4.65 \\

\multicolumn{2}{c}{w/o $f_L$} & 81.50 & 80.95 & 53.21 & 42.23 & 81.77 & 76.67 & 58.39 & 48.95 \\

\midrule
\multicolumn{2}{c}{Naive Distillation} & 83.43 & 82.52 & 54.60 & 43.81 & 80.26 & 79.13 & 57.38 & 48.92 \\
\multicolumn{2}{c}{Direct Ensemble} & 80.67 & 80.68 & 50.44 & 39.99 & 80.89 & 76.35 & 59.52 & 43.80 \\
\midrule
\multicolumn{2}{c}{D-SINK (Ours)} & \textbf{89.00} & \textbf{86.93} & \textbf{58.96} & \textbf{48.26} & \textbf{87.09} & \textbf{84.36} & \textbf{61.69} & \textbf{51.56} \\
\midrule

\multicolumn{2}{l}{\textit{D-SINK Margin vs.}} \\
\multicolumn{2}{l}{\quad IB ($f_L$)} & +16.34 & +24.74 & +24.11 & +25.45 & +16.21 & +24.04 & +14.21 & +17.75 \\
\multicolumn{2}{l}{\quad DivideMix ($f_N$)} & +8.80 & +6.99 & +11.03 & +6.61 & +6.97 & +9.63 & +6.86 & +9.61 \\
\multicolumn{2}{l}{\quad Direct Ensemble} & +8.33 & +6.25 & +8.52 & +8.27 & +6.20 & +8.01 & +2.17 & +7.76 \\
\multicolumn{2}{l}{\quad Naive Distillation} & +5.57 & +4.41 & +4.36 & +4.45 & +6.83 & +5.23 & +4.31 & +2.64 \\
\bottomrule
\end{tabular}
\end{table}

\section{Conclusion}\label{sec13}

In this paper, we tackled the challenge of learning from long-tailed noisy data by proposing a novel perspective: leveraging the combined strengths of `weak' auxiliary models, each addressing only class imbalance or label noise. We introduced {\method}, a framework that effectively distills complementary knowledge from these single-robustness models into a target model via a surrogate label allocation mechanism optimized through efficient optimal transport. Our extensive experiments demonstrated that {\method} achieves significant performance gains, thereby validating the surprising effectiveness of its core strategy—synergizing these `weak', individually focused auxiliary models to navigate the complexities of the dual-challenge scenario. This work paves the way for new approaches to complex data imperfection challenges by harmonizing simpler, specialized solutions.

\textbf{Limitations and Future Work.} While D-SINK is effective, its scope has limitations that suggest avenues for future work. The current framework is designed for multi-class classification; its optimal transport formulation enforces a sum-to-one probability constraint, making it unsuitable for multi-label problems without significant modification. Similarly, D-SINK does not explicitly address open-set noise or domain shift. Its potential resilience to these challenges is not inherent but depends entirely on the out-of-distribution detection and generalization capabilities of its auxiliary ``teacher'' models ($f_L$ and $f_N$). Extending the D-SINK framework to these important scenarios is a promising direction for future investigation.

\backmatter

\section*{Declarations}

\textbf{Funding} This work is supported by the National Key R\&D Program of China (No. 2022ZD0160702), National Natural Science Foundation of China (No. 62306178) and STCSM (No. 22DZ2229005), 111 plan (No. BP0719010).

\noindent\textbf{Conflict of interest} The authors have no financial or non-financial interests to disclose that are relevant to the content of this article.

\noindent\textbf{Ethics approval and consent to participate} Not applicable.

\noindent\textbf{Consent for publication} Not applicable.

\noindent\textbf{Data availability} All datasets used are publicly available and cited. Accompanying code for sampling and noisy label generation will be made available.

\noindent\textbf{Materials availability} Not applicable.

\noindent\textbf{Code availability} The code of this work will be available after publication. 

\noindent\textbf{Author contribution} Conceptualization: Hong F., Yao J.; Methodology: Hong F.; Experiment: Huang Y., Hong F.; Writing - original draft: Hong F., Huang Y.; Writing – review \& editing: all authors; Supervision: Yao J., Li D., Zhang Y., Wang Y.

\begin{appendices}
\section{Derivation of the Sinkhorn Algorithm }\label{secA1}
Below is a brief introduction and derivation of the Optimal Transport (OT) problem and the Sinkhorn algorithm. For clarity, the notation used herein aligns with the general Optimal Transport (OT) problem definition, which might not be consistent with the notation in the main sections of this paper.

Optimal Transport (OT) provides a mathematical framework for finding the most efficient way to transform one probability distribution into another. Given two discrete probability distributions, a source distribution $r \in \mathbb{R}_+^m$ and a target distribution $c \in \mathbb{R}_+^n$, the goal is to find a ``transport plan'' that minimizes the total transportation cost.

The problem is formally defined by Kantorovich as follows:
\begin{equation}
\min_{P \in U(r, c)} \langle P, M \rangle = \min_{P \in U(r, c)} \sum_{i,j} P_{ij}M_{ij}
\end{equation}
\begin{itemize}
    \item $M \in \mathbb{R}_+^{m \times n}$ is the \textbf{cost matrix}, where $M_{ij}$ represents the cost of moving one unit of mass from the $i$-th bin of the source to the $j$-th bin of the target.
    \item $P \in \mathbb{R}_+^{m \times n}$ is the \textbf{transport plan} or \textbf{coupling matrix}. The entry $P_{ij}$ specifies the amount of mass to be moved from the $i$-th source bin to the $j$-th target bin.
    \item $U(r, c)$ is the set of all valid transport plans, defined by the marginal constraints:
    \begin{equation*}
    U(r, c) = \{ P \in \mathbb{R}_+^{m \times n} \mid P\mathbf{1}_n = r, P^T\mathbf{1}_m = c \}
    \end{equation*}
    These constraints ensure that the total mass transported out of each source bin $i$ equals $r_i$, and the total mass transported into each target bin $j$ equals $c_j$.
\end{itemize}
Solving this linear program can be computationally intensive. A common and highly effective approach is to add an entropic regularization term, which leads to a strictly convex and differentiable problem that can be solved efficiently using the Sinkhorn algorithm.

For the entropically regularized Optimal Transport (OT) problem:
\begin{equation}
    \min_{\mathbf{P} \in U(\mathbf{r}, \mathbf{c})} \sum_{i,j} P_{ij} M_{ij} + \lambda \sum_{i,j} P_{ij} \log P_{ij},
\end{equation}
subject to $\mathbf{P1}_n = \mathbf{r}$ and $\mathbf{P}^\top\mathbf{1}_m = \mathbf{c}$, where $P_{ij} \ge 0$. The Lagrangian is:
\begin{equation}
    L(\mathbf{P}, \mathbf{f}, \mathbf{g}) = \sum_{i,j} P_{ij} M_{ij} + \lambda \sum_{i,j} P_{ij} \log P_{ij} - \sum_i f_i \left(\sum_j P_{ij} - r_i\right) - \sum_j g_j \left(\sum_i P_{ij} - c_j\right).
\end{equation}
Setting $\frac{\partial L}{\partial P_{ij}} = M_{ij} + \lambda (\log P_{ij} + 1) - f_i - g_j = 0$ yields:
\begin{equation}
    P_{ij} = \exp\left(\frac{f_i + g_j - M_{ij}}{\lambda} - 1\right).
\end{equation}
Let $K_{ij} = \exp(-M_{ij}/\lambda)$. We can write $P_{ij} = u_i K_{ij} v_j$	
  by defining scaling factors $u_i = \exp(f_i/\lambda - 1/2)$ and $v_j = \exp(g_j/\lambda - 1/2)$ (or absorbing constants differently). The key structural form is $\mathbf{P} = \text{diag}(\mathbf{u}) \mathbf{K} \text{diag}(\mathbf{v})$.

  The marginal constraints lead to:
\begin{enumerate}[leftmargin=1pt]
    \item $u_i (\mathbf{K}\mathbf{v})_i = r_i \Rightarrow \mathbf{u} = \mathbf{r} ./ (\mathbf{K}\mathbf{v})$
    \item $v_j (\mathbf{K}^T\mathbf{u})_j = c_j \Rightarrow \mathbf{v} = \mathbf{c} ./ (\mathbf{K}^T\mathbf{u})$
\end{enumerate}

\section{Hyperparameter Settings for Baselines}\label{secB1}

To ensure a fair comparison, this appendix details the hyperparameter settings for all methods. Common training parameters were kept consistent across all experiments \cref{tab:common_hyperparams}. For sensitive, method-specific parameters, we performed a grid search based on recommendations from the original papers. The search spaces and best-performing values are summarized in \cref{tab:specific_hyperparams}.

\begin{table}[h!]
\centering
\captionsetup{width=\textwidth} 
\caption{Common training hyperparameters. These foundational settings were applied to all methods, including our own, to ensure a controlled and fair comparison environment.}
\label{tab:common_hyperparams}
\begin{tabular*}{0.5\textwidth}{@{\extracolsep{\fill}}lc}
\toprule
\textbf{Parameter} & \textbf{Value} \\
\midrule
Backbone & ResNet-18 \\
Optimizer & SGD \\
Initial lr & 0.02 \\
Weight Decay & 5e-4 \\
Momentum & 0.9 \\
Total Epochs & 300 \\
Batch Size & 64 \\
\bottomrule
\end{tabular*}
\end{table}

\begin{table}[h!]
\centering
\captionsetup{width=\textwidth}
\caption{Method-specific hyperparameters and tuning details. For each key parameter, we list the search space explored via grid search and the best-performing value that was ultimately used in our main experiments.}
\label{tab:specific_hyperparams}
\begin{tabular}{lllc}
\toprule
\textbf{Method} & \textbf{Parameter Name} & \textbf{Search Space} & \textbf{Best Value Used} \\
\midrule
DivideMix & lambda\_u & \{20, 25, 30\} & 25 \\
\addlinespace
\midrule
UNICON & lambda\_c & \{0.02, 0.025, 0.03\} & 0.025 \\
\addlinespace
\midrule
LA & tau & \{0.8, 1.0, 1.2\} & 1.0 \\
\addlinespace
\midrule
LDAM & C & \{4, 5, 6\} & 5 \\
\addlinespace
\midrule
IB & start\_ib\_epoch & \{90, 100, 110\} & 100 \\
\addlinespace
\midrule
\multirow{3}{*}{TABASCO} & alpha & \{3, 4, 5\} & 4 \\
 & lambda\_u & \{20, 25, 30\} & 25 \\
 & p\_threshold & \{0.4, 0.5, 0.6\} & 0.5 \\
 \addlinespace
 \midrule
\multirow{2}{*}{FR} & lambda\_ & \{0.9, 1.0, 1.1\} & 1.0 \\
& cluster\_type & \{KNN, G2\} & KNN \\
\addlinespace
\midrule
HAR & reg\_weight & \{0.05, 0.1, 0.2\} & 0.1 \\
\addlinespace
\midrule
\multirow{2}{*}{RCAL} & beta & \{1.5, 2.0, 2.5\} & 2.0 \\
 & coff & \{0.9, 1.0, 1.1\} & 1.0 \\
\bottomrule
\end{tabular}
\end{table}

\end{appendices}

\newpage
\bibliography{reference}%

\end{document}